\documentclass[10pt,twocolumn,letterpaper]{article}

\usepackage{cvpr}              

\usepackage{graphicx}
\usepackage{amsmath}
\usepackage{amssymb}
\usepackage{booktabs}
\usepackage[accsupp]{axessibility}  

\usepackage{tabularx}
\usepackage{pifont}
\newcommand{\cmark}{\ding{51}}%
\newcommand{\xmark}{\ding{55}}%
\usepackage[pagebackref,breaklinks,colorlinks]{hyperref}
\usepackage{siunitx}

\usepackage[capitalize]{cleveref}
\crefname{section}{Sec.}{Secs.}
\Crefname{section}{Section}{Sections}
\Crefname{table}{Table}{Tables}
\crefname{table}{Tab.}{Tabs.}

\def\cvprPaperID{8} 
\def\confName{CVPR}
\def\confYear{2023}

\begin{document}











\title{Scan2LoD3: Reconstructing semantic 3D building models at LoD3 \\ using ray casting and Bayesian networks}



\author{Olaf Wysocki\textsuperscript{1 }, Yan Xia~\textsuperscript{1,2 }, Magdalena Wysocki~\textsuperscript{1 }, Eleonora Grilli~\textsuperscript{3 },\\ Ludwig Hoegner~\textsuperscript{1,4 }, Daniel Cremers~\textsuperscript{1,2 }, Uwe Stilla~\textsuperscript{1 }\\ \\
\textsuperscript{1 }Technical University of Munich, \textsuperscript{2 }University of Oxford, \textsuperscript{3 }Bruno Kessler Foundation,\\ \textsuperscript{4 }Munich University of Applied Sciences\\
{\tt\small (olaf.wysocki, yan.xia, magdalena.wysocki, ludwig.hoegner, cremers, stilla)@tum.de}\\
{\tt\small elegrilli5@gmail.com}
}
\maketitle

\begin{abstract}

Reconstructing semantic 3D building models at the level of detail (LoD) 3 is a long-standing challenge.
Unlike mesh-based models, they require watertight geometry and object-wise semantics at the façade level.
The principal challenge of such demanding semantic 3D reconstruction is reliable façade-level semantic segmentation of 3D input data.
We present a novel method, called Scan2LoD3, that accurately reconstructs semantic LoD3 building models by improving façade-level semantic 3D segmentation.
To this end, we leverage laser physics and 3D building model priors to probabilistically identify model conflicts.
These probabilistic physical conflicts propose locations of model openings: 
Their final semantics and shapes are inferred in a Bayesian network fusing multimodal probabilistic maps of conflicts, 3D point clouds, and 2D images.
To fulfill demanding LoD3 requirements, we use the estimated shapes to cut openings in 3D building priors and fit semantic 3D objects from a library of façade objects.
Extensive experiments on the TUM city campus datasets demonstrate the superior performance of the proposed Scan2LoD3 over the state-of-the-art methods in façade-level detection, semantic segmentation, and LoD3 building model reconstruction.
We believe our method can foster the development of probability-driven semantic 3D reconstruction at LoD3 since not only the high-definition reconstruction but also reconstruction confidence becomes pivotal for various applications such as autonomous driving and urban simulations.

\end{abstract}

\section{Introduction}
\label{sec:intro}

Reconstructing detailed semantic 3D building models is a fundamental challenge in both photogrammetry~\cite{HAALA2010570} and computer vision~\cite{szeliski2010computer}.
Recent developments have shown that reconstruction using 2D building footprints and aerial observations provides building models up to level of detail (LoD) 2~\cite{RoschlaubBatscheider,HAALA2010570,benchmarking3Dcitymodels}, which are characterized by complex roof shapes but display planar façades. 
Owing to their watertightness and object-oriented modeling, such models have found many applications \cite{biljeckiApplications3DCity2015} and are now ubiquitous, as exemplified by around 140 million open access building models in the United States, Switzerland, and Poland~\footnote{https://github.com/OloOcki/awesome-citygml}.

However, reconstructing façade-detailed semantic LoD3 building models remains an open challenge.
Currently, LoD3-specific façade elements, such as windows and doors, are frequently manually modeled~\cite{uggla2023future,manualLoD3seismic}; yet at-scale, automatic LoD3 reconstruction is required by numerous applications ranging from simulating flood damage~\cite{amirebrahimi2016bim}, estimating heating demand~\cite{nouvel2013citygml}, calculating façade solar potential~\cite{willenborgIntegration2018} to testing automated driving functions~\cite{schwabRequirementAnalysis3d2019}.
\begin{figure}
    \centering
    \includegraphics[width=0.8\linewidth]{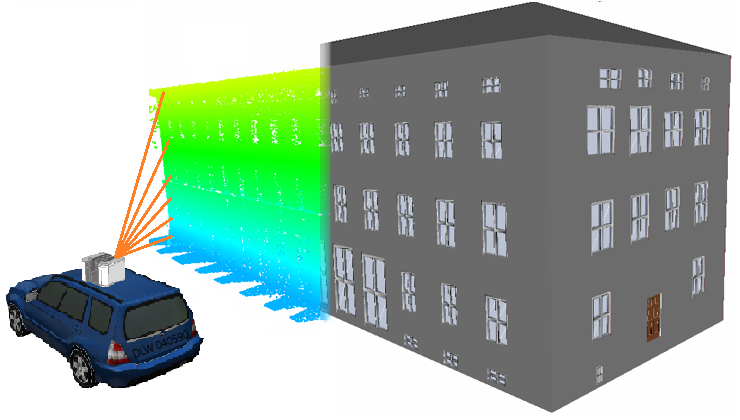}
    \caption{Scan2LoD3: Our method reconstructs detailed semantic 3D building models; Its backbone is laser rays' physics providing geometrical cues enhancing semantic segmentation accuracy.}
    \label{fig:applications}
\end{figure}

The best data source for semantic LoD3 façade modelling~\cite{xu2021towards} appears to be mobile mapping data, as the last years have witnessed a growth in mobile mapping units yielding accurate, dense, street-level image and point cloud measurements.
Yet, typically such data necessities robust, accurate, and complete semantic segmentation before it can be applied to semantic reconstruction.
In the past decade, various learning-based façade-level 3D point cloud segmentation solutions have achieved promising performance~\cite{grilli2020machine, matrone2020comparing, wysockiVisibility}. 
However, they have limited accuracy of up to 40\%~\cite{matrone2020comparing} when working on translucent (e.g., windows) and label-sparse (e.g., door) objects.
Methods based on intersections of laser rays with 3D models are used to improve the accuracy~\cite{wysockiVisibility, tuttas_reconstruction_2013}. 
However, such methods are prone to errors due to the limited semantic information~\cite{tuttas_reconstruction_2013} and field-of-view obstacles, such as window blinds~\cite{wysockiVisibility}. 
Another approaches employ images for façade segmentation and achieve high performance~\cite{riemenschneider2012irregular,liu2020deepfacade}; 
yet, their direct application for 3D façade segmentation is limited chiefly owing to the 2D representation~\cite{helmutMayerLoD3,pantoja2022generating}.

In this paper, we present a novel ray-casting-based multimodal framework for semantic LoD3 building model reconstruction named Scan2LoD3.
In contrast to previous methods, we combine multimodalties instead of relying on single modality \cite{tuttas_reconstruction_2013}; and we fuse modalities using their state probabilities, as opposed to mere binary fusion \cite{wysockiVisibility}.
The key to maintaining geometric detail is to utilize laser ray physical intersections with vector priors to find probability-quantified model conflicts in a Bayesian network, as highlighted in~\Cref{fig:applications}; we list our contributions as follows:
\begin{itemize}
    \item A probabilistic visibility analysis using mobile laser scanning (MLS) point clouds and semantic 3D building models, enabling detection of detailed conflicts by non-binary probability masks and L2 norm;
    \item A Bayesian network approach for the late fusion of multimodal probability maps enhancing 3D semantic segmentation at the façade-level;
    \item An automatic, watertight reconstruction of LoD3 models with façade elements of windows and doors compliant with the CityGML standard \cite{grogerOGCCityGeography2012};
    \item An open LoD3 reconstruction benchmark comprising LoD3 and façade-textured LoD2 building models, and façade-level semantic 3D MLS point clouds~\footnote{https://sites.google.com/view/olafwysocki/papers/scan2lod3}.  
\end{itemize}
\section{Related work}
\label{sec:relatedwork}
\begin{figure*}
    \centering
    \includegraphics[width=\linewidth]{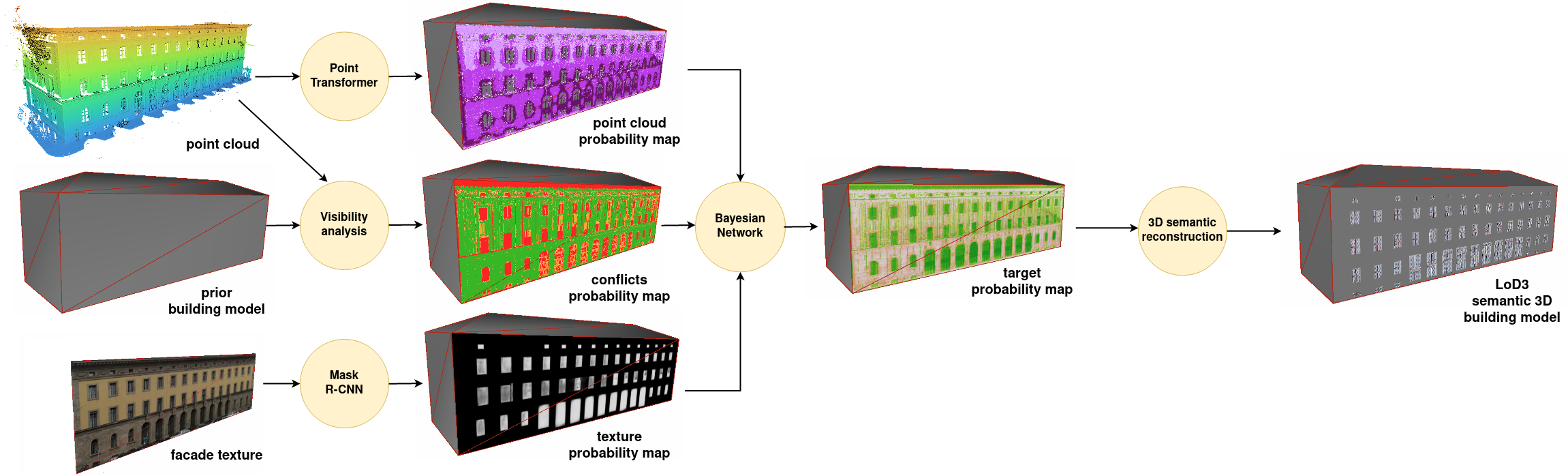}
    \caption{The workflow of the proposed Scan2LoD3 consists of three parallel branches: The first is generating the point cloud probability map based on a modified Point Transformer network (top); the second is producing a conflicts probability map from the visibility of the laser scanner in conjunction with a 3D building model (middle); and the third is using Mask-RCNN to obtain a texture probability map from 2D images. 
    We then fuse three probability maps with a Bayesian network to obtain final facade-level segmentation, enabling a CityGML-compliant LoD3 building model reconstruction.}
    \label{fig:gist}
\end{figure*}
The key to reconstructing the LoD3 building model is to achieve an accurate 3D façade segmentation.
Here, we provide insights into visibility- and learning-based methods.

{\bf Visibility analysis using ray casting and 3D models.}
In the context of 3D building models, ray casting from the sensor's origin yields deterministic information about measured, unmeasured, and unknown model parts~\cite{tuttas2015validation,meyer2022change}, but also provides geometric cues, so-called conflicts, for the façade elements reconstruction~\cite{tuttas_reconstruction_2013,wysockiVisibility,hoegner2022automatic}.
For example, Tuttas et al.~\cite{tuttas_reconstruction_2013}
exploit the fact that laser scanning rays traverse glass objects to identify building openings: 
They assume that the intersection points of rays and found building planes indicate the position of windows, which are then reconstructed by minimum bounding boxes. 
Hoegner \& Gleixner~\cite{hoegner2022automatic} pursue this idea using mobile laser scanning and, besides rays intersections, they analyze empty regions in point clouds.
Due to the methods' assumption that each visible opening is a window, they do not distinguish between other openings, such as doors or underpasses.
To overcome this issue, Wysocki et al. \cite{wysockiUnderpasses} propose the conflict classification method, which infers the semantics of ray intersections with 3D models using 2D vector maps to detect and reconstruct building underpasses.
However, conflict-based methods are prone to occlusions and are limited in identifying openings that are concealed by non-translucent objects, such as blinds. 

{ \bf Machine learning in 3D façade reconstruction.}
Early learning-based façade segmentation methods~\cite{szeliski2010computer, Tylecek13,korc-forstner-tr09-etrims,gadde2016learning,riemenschneider2012irregular} typically rely on ubiquity of 2D image façade segmentation datasets and represent façade elements as 2D objects (discussed in detail in~\cite{musialski2013survey}).
Recent works utilize well-established 2D image-based neural networks to identify façade elements in images and then project them onto 3D point clouds or their derivatives, such as 3D models~\cite{helmutMayerLoD3, pantoja2022generating, pang20223d, KadaFacades}.
However, these methods frequently assume full point cloud coverage of buildings and correctly co-referenced multiple image observations from various angles.
For example, Huang et al.~\cite{helmutMayerLoD3} propose a method employing FC-DenseNet56~\cite{jegou2017one}, trained with ortho-rectified façade images, to recognize façade openings.
The labels are projected onto LoD2 building model, which is reconstructed from a drone-based photogrammetric point cloud. 
The projected window and door labels are approximated to bounding boxes, which cut openings in LoD2 solids, thereby upgrading 3D models to LoD3. 
An alternative strategy concentrates on direct 3D façade modeling from laser scanning point clouds since MLS point clouds provide detailed and accurate depth information~\cite{xia2021vpc}.
Recently, it has been demonstrated that great advances of point-wise, learning-based methods~\cite{qi2017pointnet++,zhao2021point} are applicable in the context of 3D façade segmentation~\cite{grilli2020machine,matrone2020comparing}, where an early fusion of geometric features into DGCNN~\cite{wang2019dynamic} enhances façade segmentation accuracy.
Nevertheless, sparsely represented classes, such as windows and doors, remain challenging~\cite{matrone2020comparing}.
This issue is further exacerbated by the lack of comprehensive 3D façade-level training and validation data: to the best of our knowledge, no 3D façade-level reconstruction benchmark includes textures, point clouds, and ground-truth LoD3 models~\cite{tumfacadePaper}. 

One recent work~\cite{wysockiVisibility} pursues the idea of combining geometric features and visibility analysis.
The authors merge model conflicts and inferred semantics from a modified Point Transformer architecture~\cite{zhao2021point}.
The output is added to a 3D building model face using a projection, and respective window and door openings are 3D-modeled by 3D bounding box fitting of pre-defined models.
The method, however, is limited in reconstructing windows with partially closed blinds owing to simplified probabilities to binary masks comprising only high-probability conflicts and semantics.
Additionally, the visibility analysis concerns uncertainties using L1 distance, which generalizes L2 distance measurements, rendering it less sensitive for detailed conflicts.
\section{Methodology}
\label{sec:method}

Our Scan2LoD3 method comprises two interconnected steps: semantic 3D segmentation that yields input for semantic 3D reconstruction.
As shown in~\Cref{fig:gist}, we first generate a ray-based conflicts probability map consisting of three states (\textit{conflicted}, \textit{confirmed}, and \textit{unknown}), analyzing the visibility of the laser scanner in conjunction with 3D building models (\cref{sec:visibilityAnalysis}).
However, this map is limited to the laser field-of-view and does not provide façade-specific semantics. 
To address this limitation, we additionally introduce two probability maps derived from point clouds and images: 
The former is generated by a modified Point Transformer network~\cite{zhao2021point, wysockiVisibility} (top branch), while the latter is produced using Mask-RCNN~\cite{he2017mask} (bottom branch), as described in ~\Cref{sec:3Dsegmentation,sec:2Dsegmentation}, respectively. 
We then fuse these three probability maps via a Bayesian network, resulting in a target probability map that represents the occurrence of openings and their associated probability score (\cref{sec:bayesian}).
The opening labels yield detailed 3D opening geometries for reconstruction, which is conducted with the input 3D building model and a pre-defined 3D library of openings (\cref{sec:3dReconstruction}).
Finally, we assign the respective semantics to the reconstructed parts along with the final probability score, resulting in the CityGML-compliant LoD3 building model~\cite{grogerOGCCityGeography2012}.

\subsection{Visibility analysis concerning uncertainties}
\label{sec:visibilityAnalysis}
\begin{figure*}
     \centering
     \begin{subfigure}[b]{0.49\textwidth}
         \centering
         \includegraphics[width=\textwidth]{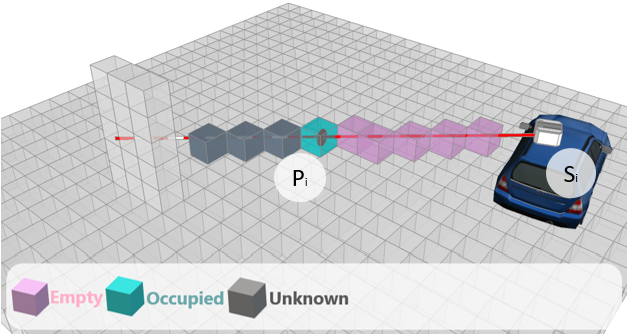}
         \caption{Ray casting of laser observations}
         \label{fig:traverse_a}
     \end{subfigure}
     \hfill
     \begin{subfigure}[b]{0.49\textwidth}
         \centering
         \includegraphics[width=\textwidth]{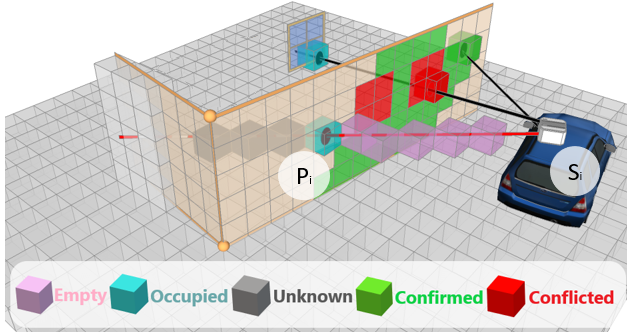}
         \caption{Rays analyzed with 3D model}
         \label{fig:traverse_b}
     \end{subfigure}
        \caption{Visibility analysis using laser scanning observations and 3D models on a voxel grid. The ray is traced from the sensor position $s_{i}$ to the hit point $p_{i}$. The voxel is: \textit{empty} if the ray traverses it; \textit{occupied} when it contains $p_{i}$; \textit{unknown} if unmeasured; \textit{confirmed} when \textit{occupied} voxel intersects with vector plane; and \textit{conflicted} when the plane intersects with an \textit{empty} voxel~\cite{wysockiVisibility}.} 
        \label{fig:traverse}
\end{figure*}

We perform ray tracing on a 3D voxel grid to determine areas that are measured by a laser scanner and analyze them with a 3D building model (\cref{fig:traverse}). 
The total grid size adapts to the input data owing to the utilized octree structure with leaves represented by 3D voxels of size $v_{s}$ dependent on the relative accuracy of the scanner.
 
As shown in~\Cref{fig:traverse_a}, the laser rays are traced from sensor position $s_{i}$, using orientation vector $r_{i}$, to hit point $p_{i} = s_{i} + r_{i}$.
Our approach leverages MLS trait of multiple laser observations~$z_{i}$ to decide upon the laser occupancy states (i.e.,~\textit{empty},~\textit{occupied}, and~\textit{unknown}) and includes the respective occupancy probability score.
The states' update mechanism uses prior probability $P(n)$, current estimate $L(n|z_{i})$, and preceding estimate $L(n|z_{1:i - 1})$ to calculate and assign the final state.
The mechanism is controlled by log-odd values~$L(n)$ along with clamping thresholds $l_{min}$ and $l_{max}$~\cite{hornung2013octomap,wysockiUnderpasses,wysockiVisibility}:
\begin{equation}
    L(n|z_{1:i}) = max(min(L(n|z_{1:i - 1}) + L(n|z_{i}), l_{max}),l_{min}) 
\end{equation}
where
\begin{equation}
     L(n) = log [\frac{P(n)}{1 - P(n)}] 
\end{equation}
%

As illustrated in~\Cref{fig:traverse_a}, in the visibility analysis process of laser observations, voxels encompassing $p_{i}$ are deemed as \textit{occupied} (light-blue), those traversed by a ray as \textit{empty} (pink), and unmeasured as \textit{unknown} (gray).
Then, as shown in~\Cref{fig:traverse_b}, we assign further voxel states by analyzing occupancy voxels and building model:
Voxels are \textit{confirmed} (green) when \textit{occupied} voxels intersect with the building surface and are \textit{conflicted} (red) when a ray traverses a building surface and reflects inside a building.
The final probability estimate, however, also concerns 3D model uncertainties.

Specifically, we address the uncertainties of global positioning accuracy of building model surfaces and of point clouds along the ray.
Let us assume that the probability distribution of global positioning accuracy of a building surface $P(A)$ is described by the Gaussian distribution \begin{math} {\mathcal{N}} (\mu_{1}, \sigma_{1}) \end{math}, where $\mu_{1}$ and $\sigma_{1}$ are the
mean and standard deviation of the Gaussian distribution.
Analogically, let us assume that the probability distribution of global positioning accuracy of a point in point cloud $P(B)$ is described by the Gaussian distribution ${\mathcal{N}} (\mu_{2}, \sigma_{2})$.
To estimate the probability of the confirmed $P_{confirmed}$ and conflicted $P_{conflicted}$ states of the voxel $V_{n}$, we use the joint probability distribution of two independent events $P(A)$ and $P(B)$: 
\begin{equation}
V_{n} = 
 \begin{Bmatrix}
  P_{confirmed}(A,B) = P(A) * P(B)\\
  P_{conflicted}(A,B) = 1 - P_{confirmed}(A,B)\\
 \end{Bmatrix} 
\end{equation}

\begin{figure}
    \centering
    \includegraphics[width=0.5\linewidth]{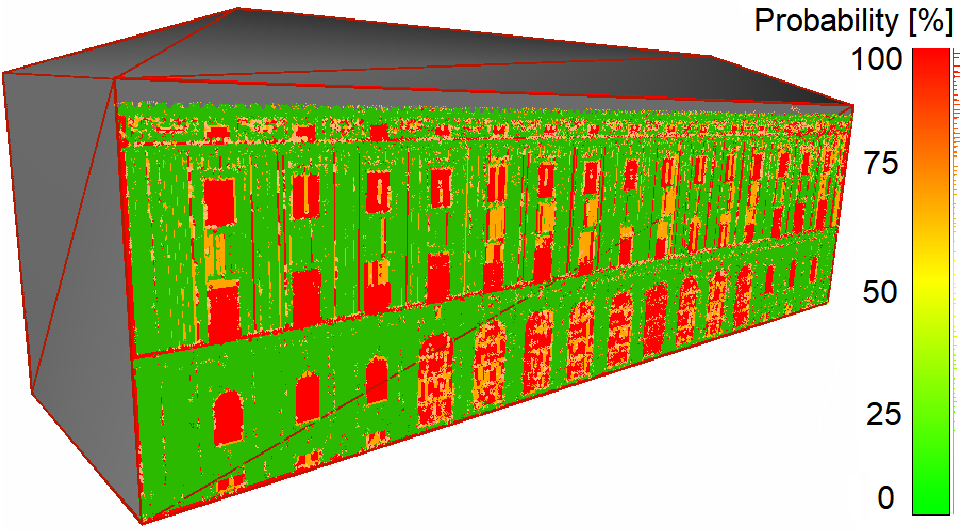}
    \caption{Exemplary~\textit{conflict probability map}: high probability pixels present high conflict probability, whereas low probability pixels show high confirmation probability.}
    \label{fig:conflictmap}
\end{figure}

We obtain a~\textit{conflicts probability map} (\cref{fig:conflictmap}) by projecting the vector-intersecting voxels to the vector plane, where the cell spacing is consistent with the voxel grid;
each pixel receives probability values of the states \textit{conflicted}, \textit{confirmed}, and \textit{unknown}, accordingly.

\subsection{3D semantic segmentation on point clouds}
\label{sec:3Dsegmentation}

We semantically segment 3D point clouds using the enhanced Point Transformer (PT) network~\cite{wysockiVisibility,zhao2021point}. 
The enhancement involves fusing geometric features at the early training stage to increase 3D façade segmentation performance~\cite{matrone2020comparing, wysockiVisibility}.
In this work, we consider seven geometric features:~\textit{height of the points, roughness, volume density, verticality, omnivariance, planarity}, and~\textit{surface variation}~\cite{jutziFeatures, grilli2020machine, wysockiVisibility}, which are calculated within an Euclidean neighborhood search radius $d_{i}$.
We define eight pertinent classes for the façade segmentation task:~\textit{arch},~\textit{column},~\textit{molding},~\textit{floor},~\textit{door},~\textit{window},~\textit{wall}, and~\textit{other}~\cite{wysockiVisibility}.

\begin{figure}[h]
    \centering
    \includegraphics[width=0.7\linewidth]{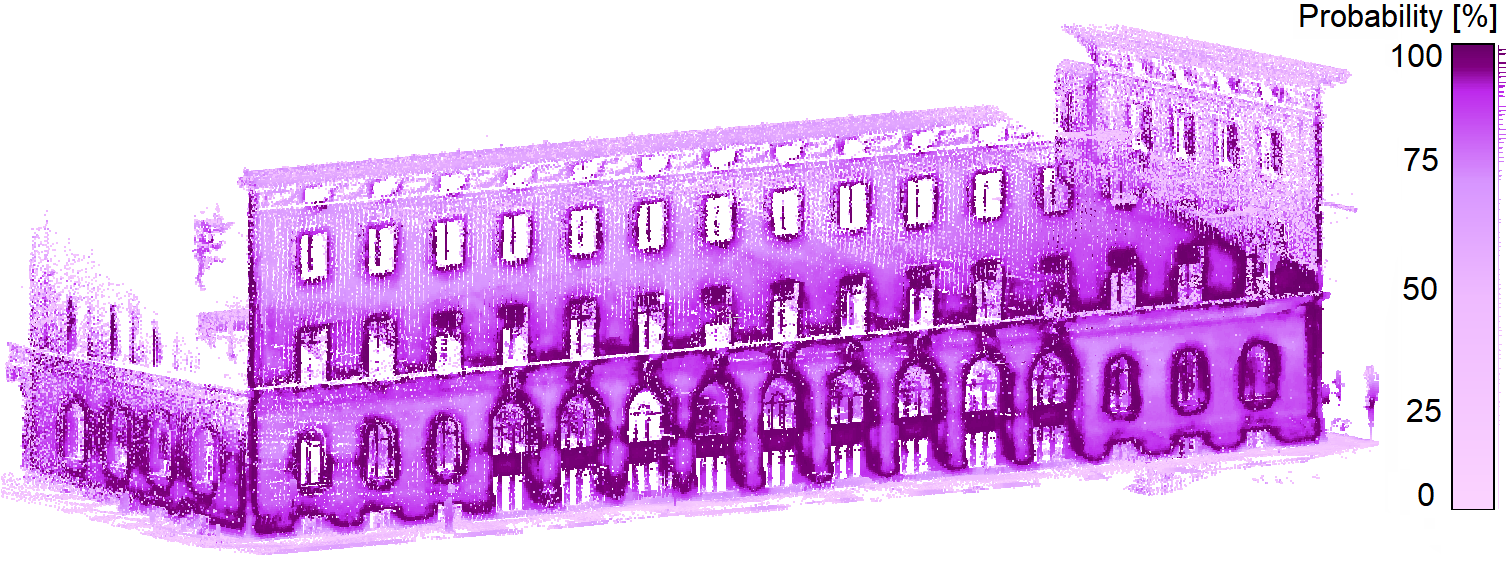}
    \caption{Exemplary results of the modified network: point cloud colors according to the probability vector of the class~\textit{window}.}
    \label{fig:probabilitiesVector}
\end{figure}

The final softmax layer of the modified PT network provides a per-point vector of probabilities of each class as an output (\cref{fig:probabilitiesVector}).
Notably, in contrast to \cite{wysockiVisibility}, we do not discard points based on a probability threshold but consider each point and its class probability score for further processing.
Finally, we create the \textit{point cloud probability map} (\cref{fig:bayesian}) by projecting the points onto the face of a building while preserving the probabilities and following the cell spacing of the \textit{conflict probability map}~(\cref{sec:visibilityAnalysis}).

\subsection{2D semantic segmentation on images}
\label{sec:2Dsegmentation}

As demonstrated by Hensel et al., 2019,~\cite{KadaFacades}, Faster R-CNN~\cite{ren2015faster} effectively identifies approximate façade openings positions. 
In our approach, we utilize Mask-RCNN~\cite{he2017mask}, which builds upon the concept of Faster R-CNN and identifies probability masks within proposed bounding boxes.
This trait allows us to obtain later a more accurate instances that are not necessarily restricted to a rectangular shape.
\begin{figure}[h]
    \centering
    \includegraphics[width=0.55\linewidth]{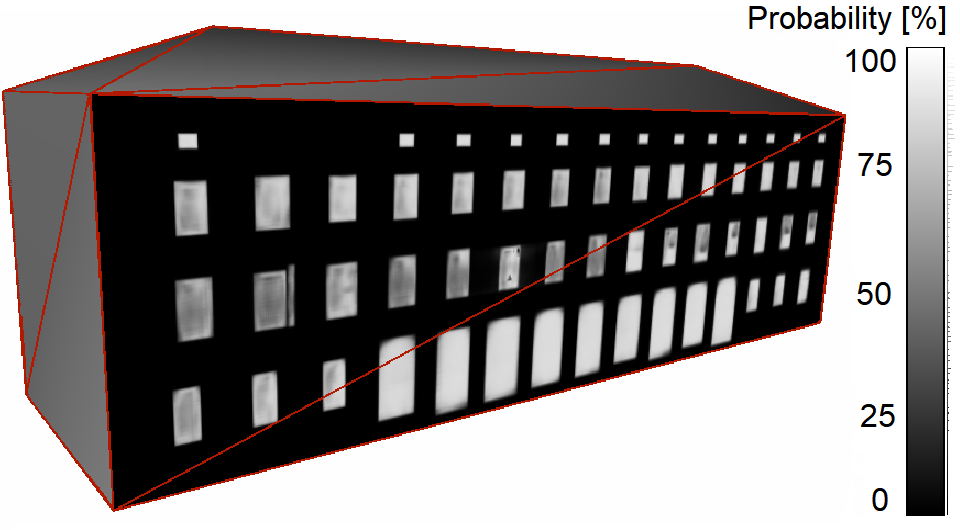}
    \caption{Exemplary~\textit{texture probability map}: high probability pixels stand for a high probability of opening.}
    \label{fig:probabilitiesMaskRCNN}
\end{figure}
For the proposed façade opening detection, we focus on two classes: windows and doors. 
Analogically to the 3D semantic segmentation stage (\cref{sec:3Dsegmentation}), we preserve the pixel-predicted probabilities.
To generate the texture probability map (\cref{fig:probabilitiesMaskRCNN}), we project the pixels and their probabilities onto the building face, aligning with the cell spacing of the other probability maps (\cref{sec:3Dsegmentation,sec:visibilityAnalysis}).

\subsection{Final segmentation with Bayesian network}
\label{sec:bayesian}

\begin{figure}
    \centering
    \includegraphics[width=0.65\linewidth]{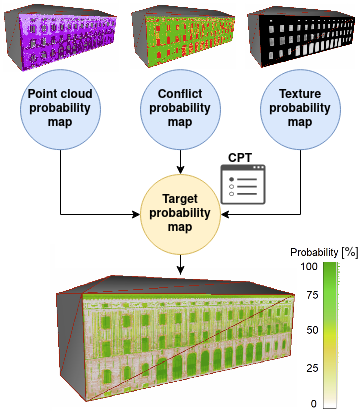}
    \caption{The Bayesian network architecture comprising three input nodes (blue), one target node (yellow), and a conditional probability table (CPT) with the assigned combinations' weights.}
    \label{fig:bayesian}
\end{figure}

To calculate the final shape, semantics, and probability score of opening instances, the multimodal probability maps are fused using a Bayesian network.
The network quantifies uncertainties and assigns weights based on evidence when calculating the target probability map. 
\Cref{fig:bayesian} shows the network architecture, including three input nodes for each probability map, to infer the probability of opening occurrence. 
The $X$ and $Y$ nodes exhibit a causal relationship, forming directed acyclic links. 
We utilize a conditional probability table (CPT) to assign weights to combinations of each node and state. 
The target node estimates two mutually exclusive states: opening and non-opening.
The probability of node $Y$ (opening space) being in the state $y$ (opening) is calculated using the marginalization process, which combines the conditional probabilities of the parent nodes' $X$ states $x$ (i.e., of point cloud probability, conflicts probability, texture probability maps) ~\cite{stritih2020online,wysockiUnderpasses}.

The probability maps serve as pieces of evidence updating the joint probability distribution $P(X, Y)$ of the compiled network. 
The inference mechanism performs the update and estimates the posterior probability distribution (PPD), which provides the states' probability~\cite{stritih2020online, wysockiUnderpasses}. 
In general, the network favors situations where there is a high probability of an opening occurring if at least two pieces of high-probability evidence co-occur; otherwise, it yields a low opening probability.
For example, a very high \textit{conflict} probability overlying high texture \textit{opening} probability and medium point cloud \textit{opening} probability should yield a high \textit{opening} probability.

As an output from the Bayesian network, we extract the high probability clusters $P_{high}$, which have a neighbor in any of the eight directions of the pixel. To distinguish between doors and windows, we compare overlying per-pixel class probabilities and select the more probable pixel class. 
The pixel-wise probability scores are then averaged per instance and kept for the final 3D model.
Since the extraction can include noisy clusters, we employ their post-processing to obtain final, noise-free opening shapes.
To this end, we apply morphological opening to reduce the effect of small distortions and weak-connected shapes.
We also calculate a modified rectangularity index~\cite{basaraner2017performance,wysockiVisibility}, on which basis we reject erroneously elongated shapes using upper $PE_{up}$ and lower $PE_{lo}$ percentiles of the index score.

\subsection{Semantic 3D reconstruction}
\label{sec:3dReconstruction}

Since it is crucial to preserve the 3D model's watertightness and its given semantics, we use the prior building solid as the basis for the modeling. 
Specifically, the openings are cut automatically in the prior model using the constructive solid geometry (CSG) difference operation: the bounding boxes of found windows and doors cut the openings in the outer boundaries of the given solid.
Then, we use these 3D cuts as matching and fitting geometries for automatically queried 3D models from a pre-defined library of LoD3 façade objects.
To ensure the watertightness and prevent self-intersections, each object is aligned with the respective face and scaled to the 3D cut shape.

We leverage the CityGML's traits to create a hierarchical semantic model structure~\cite{grogerOGCCityGeography2012}.
Specifically, the prior solid and its constituting faces preserve their unique identifiers and associated semantic classes.
The new unique identifiers are assigned to openings, which point to the respective solid's faces; 
each window and door obtain the standard~\textit{Window} and~\textit{Door} class, respectively.
As it is pivotal to preserve the final detection confidence, we also add an attribute named~\textit{confidence}, keeping the final detection confidence of the shape opening.
Ultimately, the model's LoD attribute value is upgraded to LoD3.

\section{Experiments}
\label{sec:experiments}

In this section, we describe experiments concerning the proposed Scan2LoD3 method, which necessitated acquiring existing and creating new datasets.
Within the scope of this work, we publish in the repository~\footnotemark[2]:
textured LoD2 and modeled LoD3 building models, enriched TUM-FAÇADE point clouds, implementation, and settings.
\subsection{Datasets}
To showcase the performance of Scan2LoD3, we evaluated the method on the public datasets: TUM-MLS-2016 \cite{zhu_tum-mls-2016_2020}, TUM-FAÇADE \cite{tumfacadePaper, mediatum1636761} and textured CityGML building models at LoD2~\cite{tumLoD2bavariaGov} representing the Technical University of Munich main campus in Munich, Germany.
Additionally, we used a proprietary MLS point cloud of the TUM area called MF. 
To validate the reconstruction, segmentation, and detection performance, we manually modeled a CityGML-compliant LoD3 building model~\cite{grogerOGCCityGeography2012} based on the combination of point clouds and LoD2 building model, serving as ground-truth;
the LoD2 building models were additionally textured.

{\bf The TUM-MLS-2016 dataset.} 
The point clouds in TUM-MLS-2016 were collected via obliquely mounted two Velodyne HDL-64E LiDAR sensors mounted on the Mobile Distributed Situation Awareness (MODISSA) platform. 
The entire point cloud covered an urban area with an inner and outer yard of the campus.
The inertial navigation system was supported by the real-time kinematic (RTK) correction data of the the German satellite positioning service (SAPOS), which ensured geo-referencing. 

{\bf The TUM-FAÇADE dataset.}
The TUM-FAÇADE dataset is derived from the TUM-MLS-2016 point clouds, where the former enriches the latter in 17 façade-level semantic classes.
The dataset comprises 17 annotated and 12 non-annotated façades totalling 256 million façade-level labeled and geo-referenced points. 
Within the scope of this work, we additionally annotated four of the open-access non-annotated façades. 
As discussed in \Cref{sec:3Dsegmentation}, we define seven façade classes as pertinent for the reconstruction. 
Therefore, we combined 17 TUM-FAÇADE's classes into seven by merging:
 \textit{molding} with \textit{decoration}; \textit{drainpipe} with \textit{wall}, \textit{outer ceiling surface} and \textit{stairs}; \textit{floor} with \textit{terrain} and \textit{ground surface}; \textit{other} with \textit{interior} and \textit{roof}; \textit{blinds} with \textit{window}; whereas \textit{door} remained intact.
 
{\bf The MF dataset.}
The MF point clouds were acquired at the TUM campus and covered an approximately the same area as the TUM-MLS-2016 dataset.
The point cloud was geo-referenced by proprietary mobile mapping platform, supported by the German SAPOS RTK system~\cite{mofa}.

{\bf Textured LoD2 and LoD3 semantic building models.}
 We acquired open data CityGML-compliant building priors at LoD2 from the state open access portal of Bavaria, Germany~\cite{tumLoD2bavariaGov}, which were created using 2D cadastre footprints in combination with aerial observations~\cite{RoschlaubBatscheider}; comparable results can be achieved with methods such as PolyFit~\cite{nan2017polyfit}. 
The textures were acquired manually at an approximately 45$^{\circ}$ horizontal angle using a 13MP rear camera of a Xiaomi Redmi Note 5A smartphone and projected to the respective faces: this approach simulated terrestrial acquisition of a mobile mapping unit or street view imagery where no ortho-rectifications were applied~\cite{biljeckiQualityStreetView}.
The LoD3 building model was created manually based on a combination of TUM-FAÇADE and textured LoD2 models. 
We modeled the so-called~\textit{building 23} as it has been commonly used as a validation object for various methods~\cite{tumfacadePaper,mediatum1636761, hoegner2022automatic,tuttas_reconstruction_2013,wysockiVisibility}.
The pre-defined library of openings was downloaded from the open dataset of LoD3 building models of Ingolstadt, Germany~\cite{ingolstadtLoD3}.

\subsection{Implementation details}
\label{sec:Implementation}

{ \bf Visibility analysis.} We set the size of voxels to $v_{s} = 0.1$ $m$ and initialized them with a uniform prior probability of $P = 0.5$ to perform the ray casting on an efficient octree structure~\cite{hornung2013octomap}; we used the standard~\cite{hornung2013octomap,tuttas2015validation,wysockiVisibility} clamping and log-odd values. 
The uncertainty of building models and point clouds was assigned considering their reported global positioning accuracy.
As such, the parameters of building models were set to $\mu_{1} = 0$ and $\sigma_{1} = 3$, while for the TUM-MLS-2016 and MF point clouds were set to $\mu_{2} = 0$, $\sigma_{2} = 2.85$ and to $\mu_{2} = 0$, $\sigma_{2} = 1.4$, respectively.

{\bf Semantic segmentation.} 
For the modified Point Transformer data pre-processing, we followed \cite{wysockiVisibility} and removed redundant points within a $5$ $cm$ radius, which resulted in 10 million points; 
the point cloud was split into 70\% training and 30\% validation subsets. 
We chose the optimal geometric features search radius $d_{i}$ following \cite{grilli2019geometric,wysockiVisibility}: 
As for the features \textit{roughness, volume density, omnivariance, planarity}, and \textit{surface variation} the radius was set to $d_{i} = 0.8$ $m$; 
whereas for \textit{verticality} to $d_{i} = 0.4$ $m$.
For the image segmentation, we deployed a pre-trained Mask-RCNN on the COCO dataset~\cite{lin2014microsoft}.
The inference was fine-tuned with 378 base images of the CMP façade database~\cite{Tylecek13}, where we selected two classes for training:~\textit{door} and~\textit{window} including~\textit{blinds}.
As $P_{high}$ pixels in the Bayesian network, we deemed values higher than $P_{high} = 0.7$.
To reject outliers, we fixed the modified rectangularity percentiles to  $PE_{up} = 95$ and $PE_{low} = 5$.

\subsection{Results and Discussion}
\label{sec:results}

{ \bf Detection rate.}
The methods of Hoegner \& Gleixner, 2022,~\cite{hoegner2022automatic} and Wysocki et al., 2022~\cite{wysockiVisibility} were both tested on the three façades of the~\textit{building 23} at the TUM campus using the TUM-MLS-2016 data; thus we validated the detection accuracy using the same setup and our manually modeled LoD3 building (\cref{tab:mofaDetection}).
To show the ratio of the detection rate to the laser-covered rate, we introduced metrics for all existing façade openings (AO) and only laser-measured façade openings (MO).

\begin{figure*}
    \centering
    \includegraphics[width=0.85\linewidth]{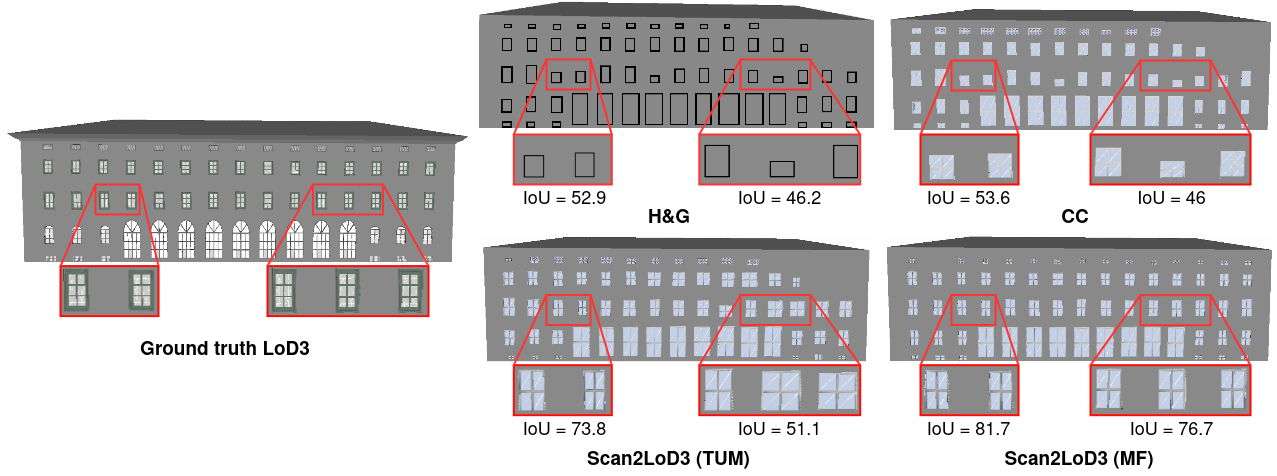}
    \caption{Comparison of different reconstruction results for the façade A: Our method reconstructs complete window shapes despite the presence of window blinds (red boxes).}
    \label{fig:reconstructionComparison}
\end{figure*}

Our multimodal fusion enabled a higher detection rate and still maintained a low false alarm rate.
If compared to the Hoegner \& Gleixner (H\&G)~\cite{hoegner2022automatic} and CC~\cite{wysockiVisibility} methods, Scan2LoD3 achieved higher detection rate on the TUM dataset by 10\% and 6\%, respectively (\cref{tab:mofaDetection} and~\cref{fig:reconstructionComparison}).
\begin{table}[h]
\centering
\footnotesize
\setlength\tabcolsep{1.8pt}
\begin{tabular}{lcccccccccccccccc}
\toprule
& \multicolumn{4}{c}{H\&G~\cite{tuttas_reconstruction_2013}} & \multicolumn{4}{c}{CC~\cite{wysockiVisibility}} & \multicolumn{4}{c}{Scan2LoD3} & \multicolumn{4}{c}{Scan2LoD3}  \\
& \multicolumn{4}{c}{} & \multicolumn{4}{c}{} & \multicolumn{4}{c}{(TUM)} & \multicolumn{4}{c}{(MF)}  \\
\cmidrule(lr){2-5} \cmidrule(lr){6-9} \cmidrule(lr){10-13} \cmidrule(lr){14-17}
 & A & B & C & Tot & A & B & C & Tot & A & B & C & Tot & A & B & C & Tot \\
 AO & 66 & 17 & 20 & \textbf{103} & 66 & 17 & 20 & \textbf{103} & 66 & 17 & 20 & \textbf{103} & 66 & 17 & 20 & \textbf{103}\\
\midrule
			 MO & 60 & 17 & 10 & \textbf{87} & 60 & 17 & 12 & \textbf{87} & 60 & 17 & 12 & \textbf{89} & 66 & 12 & 18 & \textbf{96} \\
			 D & 60 & 15 & 4 & \textbf{75} & 60 & 15 & 6 & \textbf{81} & 60 & 16 & 11 & \textbf{87} & 65 & 16 & 16 & \textbf{97} \\
			 TP & 60 & 12 & 4 & \textbf{76} & 60 & 15 & 5 & \textbf{80} & 60 & 16 & 11 & \textbf{87} & 65 & 14 & 15 & \textbf{94}\\
			 FP & 0 & 3 & 0 & \textbf{3} & 0 & 0 & 1 & \textbf{1} & 0 & 0 & 0 & \textbf{0} & 0 & 0 & 1 & \textbf{3}\\
			 FN & 6 & 5 & 16 & \textbf{27} & 6 & 2 & 15 & \textbf{23} & 6 & 1 & 9 & \textbf{16} & 1 & 3 & 5 & \textbf{9} \\\hline
DA & 91 & 71 & 20 & \textbf{74} & 91 & 88 & 25 & \textbf{78} & 91 & 94 & 55 & \textbf{84} & 98 & 82 & 75 & \textbf{91} \\
FA & 0 & 0 & 0 & \textbf{4} & 0 & 0 & 17 & \textbf{1} & 0 & 0 & 0 & \textbf{0} & 0 & 12 & 6 & \textbf{3} \\
DM & 100 & 71 & 40 & \textbf{87} & 100 & 88 & 42 & \textbf{90} & 100 & 94 & 92 & \textbf{98} & 98 & 117 & 83 & \textbf{98}\\
\bottomrule
\end{tabular}
    \caption{Detection rate for all openings (DA) and laser-measured openings (DM) and the respective false alarm rate (FA) for façades A, B, and C (AO = all openings, MO = laser-measured openings, D = detections, TP = true positives, FP = false positives, FN = false negatives).}
    \label{tab:mofaDetection}
\end{table} 
The MF map provided more accurate results (i.e., 91\% of all openings correctly detected) owing to higher point cloud global accuracy and complete façade A coverage; also other maps complemented the MF's laser-observed openings, as exemplified by façade B (\cref{tab:mofaDetection}).

{ \bf Semantic segmentation.}
\begin{figure*}
    \centering
    \includegraphics[width=0.85\linewidth]{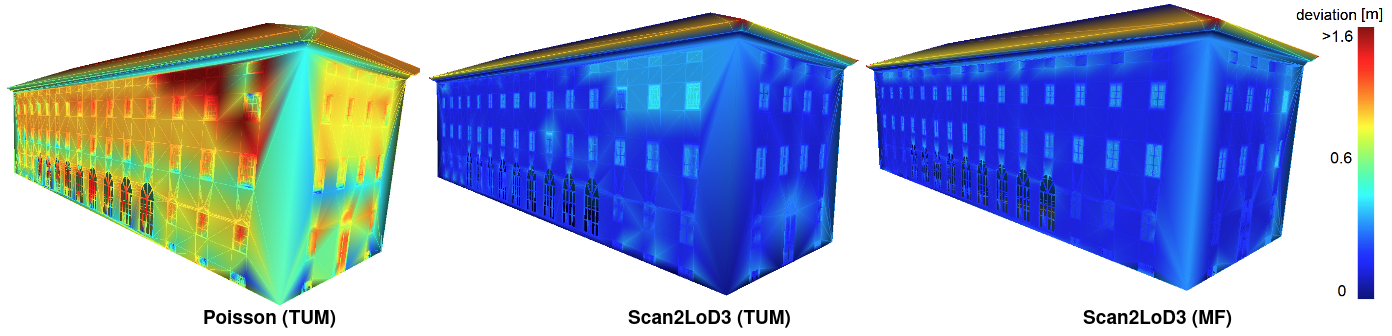}
    \caption{Comparison of the Poisson to our reconstruction approach: Deviations are projected onto the ground-truth LoD3 model.}
    \label{fig:reconstructionPoisson}
\end{figure*}
To measure the accuracy of the segmentation, we selected the median per-instance intersection over union (IoU) metric for all openings of \textit{building 23} (\cref{tab:segmentationIoU,fig:reconstructionComparison}).
This setup enabled us the comparison to the introduced modified Point Transformer network (Pt+Ft.) working only on point clouds; Mask-RCNN (M-RCNN) using only images~\cite{he2017mask}; method using ray-casting and binary point cloud masks (CC)~\cite{wysockiVisibility}; our method fusing three maps (i.e., conflicts, point clouds, images), once with the TUM-MLS-2016 conflict map (TUM) and on the higher accuracy conflict map of MF (MF).

Our experiments corroborate that, in contrast to the tested methods, our proposed solution identifies even closed openings, their full shapes, and reaches higher accuracy (\cref{tab:segmentationIoU} and~\cref{fig:reconstructionComparison}).
This fact enabled the whole-shape reconstruction of, for example, covered by blinds windows, which resulted in up to 20\% higher IoU on the TUM-MLS-2016 dataset (red boxes,~\cref{fig:reconstructionComparison}).
\begin{table}[h]
\centering
\footnotesize
\setlength\tabcolsep{4pt}
\begin{tabular}{lcccc}
\toprule 
& \multicolumn{4}{c}{median IoU $\uparrow$}\\
\cmidrule(l){2-5}
Façade & A & B & C & Total \\
Openings & 66 & 17 & 20 & 103 \\
\midrule
PT+Ft. \cite{zhao2021point}  & 7.3 & 4.6 & 3.7 & 7.3 \\
M-RCNN \cite{he2017mask}  & 63.7 & 47.4 & 38.6 & 58.4 \\
CC \cite{wysockiVisibility}  & 66.5 & 56.4 & \textbf{53.2} & 60.6 \\
Scan2LoD3 (TUM)  & 63.9 & 52.9 & 38 & 62.1 \\
Scan2LoD3 (MF)  & \textbf{78.4} & \textbf{62.3} & 40.6 & \textbf{76.2} \\
\bottomrule
\end{tabular}
	\caption{Comparison of opening segmentation using only: 3D point clouds (Pt+Ft.), images (M-RCNN), binary masks (CC), and our method with TUM and MF conflict maps.}
\label{tab:segmentationIoU}
\end{table}
Similarly to the detection results, the accuracy of laser measurements significantly influenced the IoU results: 
Our method tended to overestimate opening shapes on the TUM point cloud, whereas on MF the shapes were approximately 14\% more accurate.
On the other hand, Scan2LoD3 was sensitive to poor segmentation results (façade C,~\cref{tab:segmentationIoU}).

{ \bf 3D reconstruction.}
We measured the accuracy of reconstruction by comparing our method using the TUM-FAÇADE data to the well-established and mesh-oriented Poisson reconstruction~\cite{kazhdan2006poisson} and to the second-best-IoU performing CC method (\cref{fig:reconstructionComparison,fig:reconstructionPoisson,tab:Hausdorff}). 
To highlight the influence of point cloud accuracy, we also added the results for MF point clouds.
\begin{table}[htb]
    \footnotesize
    \centering
    \begin{tabular}{l ccc} 
	\toprule
	Method  & \multicolumn{2}{c}{vs. GT LoD3 $\downarrow$} \\
	\cmidrule(lr){2-3}
    & $\mu$ & RMS & WT\\
    \midrule
    Poisson (TUM)~\cite{kazhdan2006poisson} & 0.35 & 0.54 & \xmark \\
    CC~\cite{wysockiVisibility} & 0.31 & 0.34 & \cmark \\
    Scan2LoD3 (TUM) & 0.23 & 0.26 & \cmark \\
    Scan2LoD3 (MF)  & \textbf{0.13} & \textbf{0.25} & \cmark \\
    \bottomrule
    \end{tabular}
    \caption{Comparison of mesh-based Poisson, building-prior-driven CC, and our proposed method using the ground-truth LoD3 model and measuring watertightness (WT).}
    \label{tab:Hausdorff}
\end{table}
As shown in~\Cref{tab:Hausdorff} and in \Cref{fig:reconstructionPoisson}, the 3D building priors provided more accurate reconstruction results than the standard Poisson reconstruction (i.e., RMS lower by 52\%);
the former also achieved the watertightness.
Among the prior-driven methods, the improvement related to higher detection rate and IoU was noticeable: Scan2LoD3 had lower mean and RMS scores by up to 26\% and 24\%, respectively, compared to CC (\cref{tab:Hausdorff}). 
It is worth noting that the eaves were incorrectly reconstructed in any of the presented methods. 

\section{Conclusions}
\label{sec:conclusion}
In this paper, we introduce Scan2LoD3, a multimodal probabilistic fusion method for the high-detail semantic 3D building reconstruction.
Our work has led us to the conclusion that the multimodal probabilistic fusion can maximize the advantages of ray-casting- and learning-based methods for the LoD3 reconstruction.
The findings of this study indicate that while joining images, point clouds, and model conflicts, a Bayesian network reveals a very high-level detection rate (i.e., 91\%); and robustness as the false alarm rate is negligible (i.e., 3\%). 
Crucially, our method segments and reconstructs complete opening shapes, even when closed by blinds, which can provide up to around 76\% shape accuracy.
By such detection and segmentation, we minimize the final reconstruction deviations by 54\% and 24\% when compared to mesh-based and other prior-driven methods, respectively.
Such method's characteristics are of great importance for applications necessitating object-oriented semantics, high robustness, and completeness, such as automated driving testing~\cite{schwabRequirementAnalysis3d2019} or façade solar potential analysis \cite{willenborgIntegration2018}, among others \cite{palliwal20213d, wong2020mapping}.
Furthermore, an upshot of keeping reconstruction confidence score can be pivotal for confidence-based navigation algorithms, such as in autonomous cars~\cite{wong2020mapping,albrecht2022investigation,SesterZou}.
It is worth noting that our method focuses on upgrading facades to LoD3; refining roofs to LoD3 would require additional, airborne data.

As the late fusion results so far have been very encouraging and do not require any training data, we deem Bayesian networks suitable for the task.
Future work will concentrate on comparing the Bayesian network's generalization capabilities to deep neural networks, which, however, require extensive training data.
Moreover, we expect the method's performance to be comparable on similar architecture styles; considering selected classes and small sample size. 
To tackle these issues, we plan to extend our open library of textured LoD2 and LoD3 models to foster the methods' development.
%

{\bf Acknowledgments}
This work was supported by the Bavarian State Ministry for Economic Affairs, Regional Development and Energy within the framework of the IuK Bayern project \textit{MoFa3D - Mobile Erfassung von Fassaden mittels 3D Punktwolken}, Grant No.\ IUK643/001.
Moreover, the work was conducted within the framework of the {Leonhard Obermeyer Center} at the Technical University of Munich (TUM).
 
{\small
\bibliographystyle{ieee_fullname}
\bibliography{egbib}
}

\end{document}